\documentclass[lettersize,journal]{IEEEtran}
\usepackage{amsmath,amsfonts}
\usepackage{algorithm}
\usepackage{algorithmic}
\usepackage{array}
\usepackage{titlesec}
\usepackage[caption=false,font=normalsize,labelfont=sf,textfont=sf]{subfig}
\usepackage{amsmath}
\usepackage{textcomp}
\usepackage{stfloats}
\usepackage{url}
\usepackage{verbatim}
\usepackage{graphicx}
\usepackage{cite}
\usepackage{multirow}
\usepackage{hyperref}
\usepackage{tikz}
\usetikzlibrary{positioning}
\usepackage{physics}
\usepackage{tikz}
\usepackage{mathdots}
\usepackage{yhmath}
\usepackage{cancel}
\usepackage{color}
\usepackage{siunitx}
\usepackage{multirow}
\usepackage{amssymb}
\usepackage{gensymb}
\usepackage{tabularx}
\usepackage{extarrows}
\usepackage{booktabs}
\usetikzlibrary{fadings}
\usetikzlibrary{patterns}
\usetikzlibrary{shadows.blur}
\usetikzlibrary{shapes}
\usepackage{diagbox}
\usepackage{bbding}
\usepackage{amsthm}
\usepackage{float}

\usepackage{setspace}
\usepackage{geometry}
\usepackage{graphicx}
\usepackage{physics}
\usepackage{amsmath}
\usepackage{tikz}
\usepackage{mathdots}
\usepackage{yhmath}
\usepackage{cancel}
\usepackage{color}
\usepackage{siunitx}
\usepackage{array}
\usepackage{multirow}
\usepackage{amssymb}
\usepackage{gensymb}
\usepackage{tabularx}
\usepackage{extarrows}
\usepackage{booktabs}
\usetikzlibrary{fadings}
\usetikzlibrary{patterns}
\usetikzlibrary{shadows.blur}
\usetikzlibrary{shapes}
\usepackage{diagbox}
\usepackage{bbding}
\usepackage{amsthm}
\usepackage{float}
\usepackage{bm}
\usepackage{pdflscape}
\usepackage{rotating}
\usepackage{booktabs}

\begin{document}

\title{Adversarial Domain Adaptation for Cross-user Activity Recognition Using Diffusion-based Noise-centred Learning}

\author{
Xiaozhou Ye
\thanks{Xiaozhou Ye is with the Department of Electrical, Computer and Software Engineering, The University of Auckland, New Zealand (e-mail: xye685@aucklanduni.ac.nz)}
\href{https://orcid.org/0000-0002-9725-1548}{\includegraphics[scale=0.01]{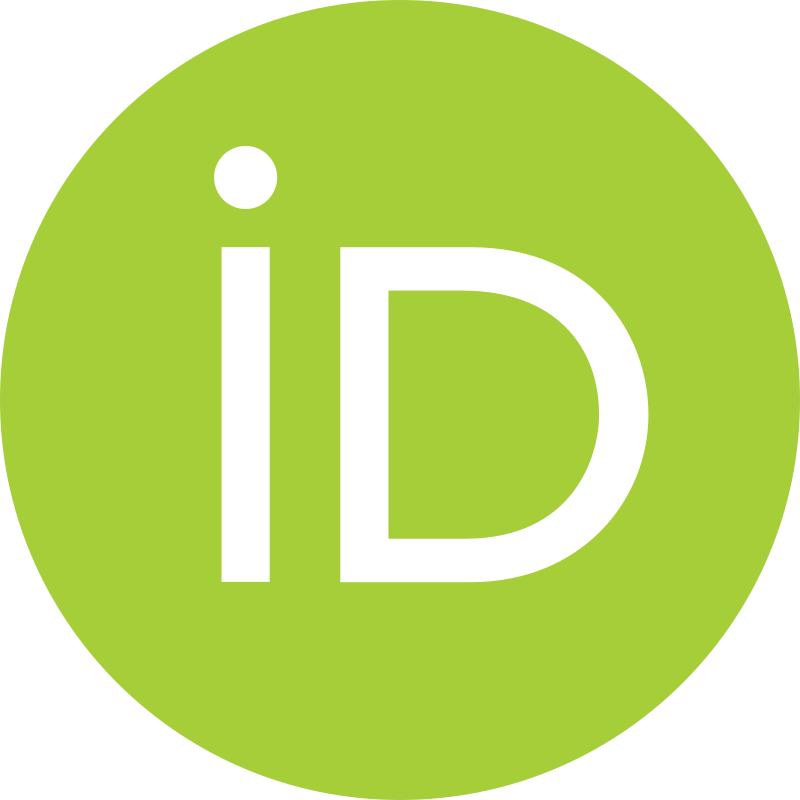}}
\and,
Kevin I-Kai Wang
\thanks{Kevin I-Kai Wang is with the Department of Electrical, Computer and Software Engineering, The University of Auckland, New Zealand (e-mail: kevin.wang@auckland.ac.nz)}
\href{https://orcid.org/0000-0001-8450-2558}{\includegraphics[scale=0.01]{ORCID_iD.png}}}

\maketitle

\begin{abstract}
Human Activity Recognition (HAR) plays a crucial role in various applications such as human-computer interaction and healthcare monitoring. However, challenges persist in HAR models due to the data distribution differences between training and real-world data distributions, particularly evident in cross-user scenarios. This paper introduces a novel framework, termed Diffusion-based Noise-centered Adversarial Learning Domain Adaptation (DNA-DA), designed to address these challenges by leveraging generative diffusion modeling and adversarial learning domain adaptation techniques. Traditional HAR models often struggle with the diversity of user behaviors and sensor data distributions. DNA-DA innovatively integrates the inherent noise within diffusion models, harnessing its latent information to enhance domain adaptation. Specifically, the framework transforms noise into a critical carrier of activity and domain class information, facilitating robust classification across different user domains. Experimental evaluations demonstrate the effectiveness of DNA-DA in improving HAR model performance across different users, surpassing traditional domain adaptation methods. The framework not only mitigates distribution mismatches but also enhances data quality through noise-based denoising techniques.
\end{abstract}

\section{Introduction}

Human Activity Recognition (HAR) is a critical component of human-computer interaction, ubiquitous computing, and the Internet of Things. The primary objective of HAR is to accurately classify human activities using sensor data and contextual information \cite{dang2020sensor}. Despite extensive research efforts, developing an effective HAR model remains challenging. A significant issue in current HAR methodologies, particularly those dealing with time series sensor data, is the presumption that training and testing data originate from identical distributions. This assumption is frequently invalidated by the diversity of real-world data and the occurrence of out-of-distribution ($\displaystyle o.o.d. $) instances \cite{chen2021deep}. This discrepancy, caused by factors such as sensor heterogeneity \cite{xing2018enabling}, evolving data patterns \cite{lu2018learning}, varying sensor layouts \cite{rokni2018autonomous}, and individual behavioral differences \cite{xiaozhou2023temporaloptimal}, severely impairs the generalizability of HAR models. Our research specifically addresses the $\displaystyle o.o.d. $ challenge in sensor-based HAR, focusing on the cross-user HAR problem arising from individual behavioral variances, commonly encountered in healthcare applications.

Transfer learning and domain adaptation \cite{farahani2021brief} present promising solutions for bridging the data distribution gap between different domains. In cross-user Human Activity Recognition (HAR), these approaches aim to leverage knowledge from a source domain (e.g., activities of one group of users) to enhance performance in a target domain (e.g., activities of a different group of users). Techniques such as Maximum Mean Discrepancy \cite{yan2017mind} and Optimal Transport \cite{xiaozhou2023temporaloptimal} are utilized to align domain features, while more advanced methods like Deep Generative Domain Adaptation \cite{ye2024deep} and Deep Conditional Adaptation Network \cite{ge2023unsupervised} focus on deeper, non-linear alignments. Recent advancements in generative modeling have incorporated diffusion models into domain adaptation tasks, showcasing their potential to bridge the gap between source and target domains. These models facilitate smooth, controlled transitions while preserving critical data features, as demonstrated by the Domain-Adaptive Diffusion method \cite{peng2024unsupervised}. Moreover, their capacity to generate high-fidelity samples enhances accurate domain translation and feature adaptation, as evidenced by the Data Augmentation with Diffusion Models \cite{benigmim2023one}. Moreover, IoT sensor data such as human activity recognition data often contains significant noise due to various environmental and operational factors. Diffusion models, which simulate the process of information spreading or diffusing through a system, are highly effective at noise reduction. They iteratively refine the sensor data, making it smoother and more accurate by filtering out random fluctuations and disturbances. This denoising capability enhances the quality of the data, leading to more reliable activity recognition \cite{zuo2023unsupervised}. By applying diffusion models, the integrity and quality of sensor data can be significantly improved for better domain adaptation. 

\begin{figure}[h!]
\centering
\includegraphics[width=1\columnwidth]{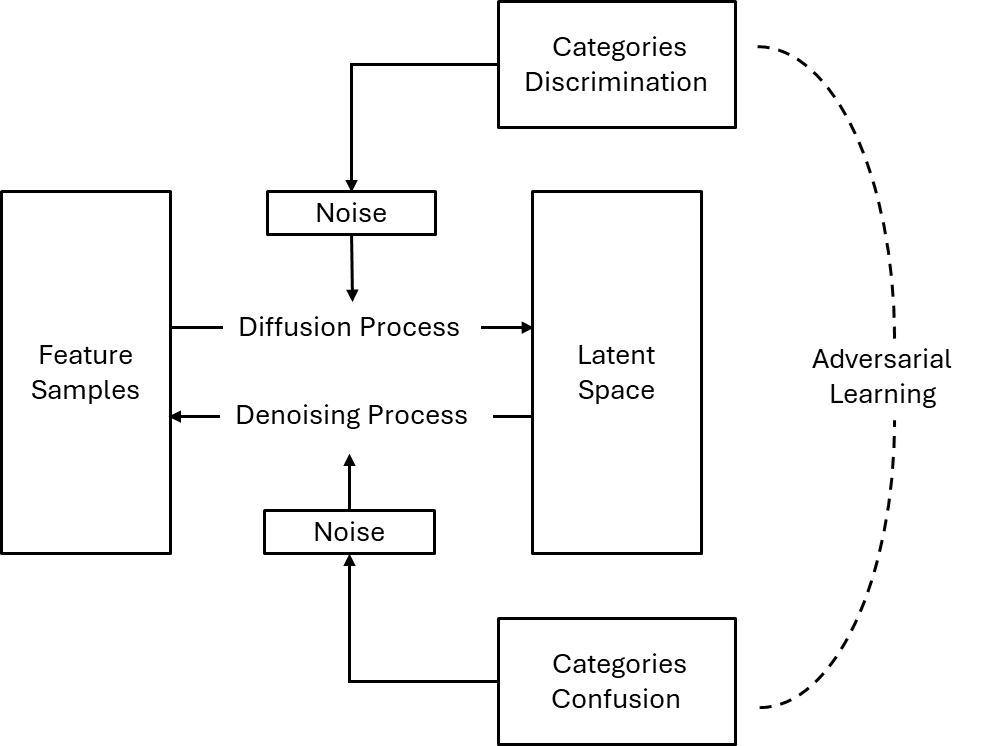}
\caption{The design of DNA-DA.\label{DNA-DA}}
\end{figure}

However, existing diffusion-based domain adaptation methods often neglect the crucial role of noise in improving domain adaptation, which can carry valuable information in both the forward diffusion process and the reverse diffusion denoising process. To address these overlooked aspects, we propose a novel approach that harnesses the potential of noise within diffusion models for domain adaptation, termed Diffusion-based Noise-centered Adversarial Learning Domain Adaptation (DNA-DA). Our method integrates noise-centered learning into the diffusion process, enabling the model to develop robust activity classification models that generalize effectively across different user domains by enhancing the utilization of noise. This method transforms the forward diffusion and reverse denoising processes into adversarial learning domain adaptation phases, aligning feature distributions between source and target domains and improving the model's ability to adapt to individual behavioral differences.

Specially, in the forward diffusion process as shown in Figure~\ref{DNA-DA}, the data is transformed into a distribution resembling pure noise. However, the noise itself is underutilized. By embedding activity and domain class information into the noise, we enable it to follow different Gaussian distributions based on these conditional inputs, thus enriching the noise with additional classification information. During the reverse diffusion denoising process, the noise is closely linked to the features in the classification task, containing valuable information about the classification labels. By accurately predicting the reverse noise, the model can extract and leverage this embedded information within noise. Simultaneously, we assign an additional task to the noise, applying reverse gradient learning techniques to obfuscate the model's ability to distinguish between activity and domain classes.

Thus, the forward diffusion and reverse denoising processes are designed as adversarial learning domain adaptation. In the forward phase, the objective is to effectively distinguish between user and activity classes, acting as a discriminator and ensuring the model can identify unique characteristics of different users and their activities. Conversely, the reverse phase aims to obscure the data distribution related to classification and users, functioning as a generator. This adversarial interaction helps the model robustly align feature distributions across domains, reducing discrepancies and enhancing generalization capabilities.

In summary, the contributions of our work are as follows:

\begin{enumerate}
\item \textbf{Novel Deep Domain Adaptation Framework}: We propose a novel deep domain adaptation framework, DNA-DA, specifically designed for cross-user HAR. This framework integrates the concepts of adversarial learning domain adaptation and generative modeling in a unique manner to exploit the knowledge embedded in the diffusion model for better domain adaptation.

\item \textbf{Innovative Generative Model Architecture}: We introduce an innovative generative model architecture, incorporating unique constraint conditions and network structures that facilitate data distribution alignment between source and target domains. This enhances the generalization capability of the diffusion model and its classification performance.

\item \textbf{Noise-centered Adversarial Learning Domain Adaptation}: We develop a unique noise-centered adversarial learning domain adaptation mechanism that transforms the noise in both forward and reverse diffusion processes into critical information carriers for classification tasks. By integrating activity and domain class information into the noise, we enable the model to extract and leverage embedded information, significantly enhancing the robustness and adaptability of HAR models across different user domains.
\end{enumerate}

The paper is organized as follows: it starts with a review of related work in sensor-based cross-user HAR and state-of-the-art methods in transfer learning and domain adaptation. Next, it delves into the specifics of the DNA-DA method, detailing its design principles and emphasis on noise-centered adversarial domain adaptation. Following this, we describe our experimental setup and compare DNA-DA with existing methods, highlighting its effectiveness in leveraging noise information within diffusion models. The paper concludes with a discussion of our findings and suggestions for future research directions in this field. The DNA-DA implementation is available at https://github.com/helloworld1973/DNA-DA.

\section{Related Work}

\subsection{Sensor-based Cross-user Activity Recognition}

Sensor-based HAR has emerged as a key component in the field of ubiquitous computing, providing deep insights into human behaviors by analyzing sensor data and contextual clues. This technology employs a variety of sensor types, each suited to different contexts, and can be grouped into five primary categories: smartphones and wearables, ambient sensors, device-free sensing, vision-based systems, and miscellaneous sensor types, as indicated in several studies \cite{chen2021deep}\cite{roche2021multimodal}. Our research specifically focuses on the use of wearable sensors for HAR.

In machine learning, the problem of HAR using sensors is viewed as a challenge in time series classification \cite{yadav2021review}. Several classification techniques have been suggested, including Bayesian networks, Support Vector Machines, and Hidden Markov Models. Additionally, deep learning models have shown impressive results across many tasks. In the context of HAR, these deep learning methods \cite{sezavar2024dcapsnet} can automatically learn to identify and extract relevant features from extensive datasets. However, these methods typically rely on the assumption that both training and testing data share the same distribution, conforming to the independent and identically distributed ($\displaystyle i.i.d.$) assumption \cite{chen2021deep}. Yet, this assumption is often violated in practical applications, where training and testing data sets usually differ significantly, falling into what is known as out-of-distribution ($\displaystyle o.o.d.$) scenarios \cite{lu2022out}. Our study explores the challenges of sensor-based HAR in these $\displaystyle o.o.d.$ conditions.

The issue of $\displaystyle o.o.d.$ in sensor-based HAR presents itself in multiple ways. Firstly, there could be discrepancies in data from different sensors, which might differ in format and distribution due to varying sensor types, platforms, manufacturers, and modalities \cite{xing2018enabling}. Secondly, data patterns can change over time—a phenomenon known as concept drift \cite{lu2018learning}. This might be observed in modifications to walking patterns due to health changes. Thirdly, there are substantial variations in behavior among individuals \cite{xiaozhou2023temporaloptimal}, such as differences in walking speeds. Finally, the specific placement of sensors on different body parts can affect data distributions \cite{rokni2018autonomous}. Our research primarily targets the challenges posed by individual behavioral variations in these $\displaystyle o.o.d.$ conditions in HAR.

\subsection{Transfer Learning and Domain Adaptation}
Transfer learning and domain adaptation have been widely studied to improve model generalization across different domains. Methods like adversarial training, feature alignment, and fine-tuning have shown promise in reducing domain shifts.

This paper \cite{han2022learning} introduces a method for identifying transferable parameters in deep unsupervised domain adaptation (UDA) networks. It focuses on optimizing the source hypothesis by including an unsupervised objective function to ensure learning of domain-invariant information. The method determines important parameters using gradient flows and weights, which are crucial for learning domain-invariant features. This work \cite{wang2022cross} presents a cross-domain contrastive learning framework for UDA, using contrastive loss to align the distributions of source and target domains. By bringing similar instances closer and pushing dissimilar ones apart, the framework learns domain-invariant representations. This paper \cite{yang2023tvt} explores the use of Vision Transformers for UDA, proposing the Transferable Vision Transformer (TVT). TVT uses self-attention mechanisms to learn domain-invariant features, enhancing generalization across domains. Extensive experiments show that TVT outperforms traditional convolutional neural networks in domain adaptation tasks. The authors \cite{dai2022graph} propose AdaGCN, a graph transfer learning framework based on adversarial domain adaptation with graph convolutional networks (GCNs). AdaGCN combines semi-supervised learning for discriminative node representations with adversarial learning for domain invariance.

Recent advancements in generative modeling have led to the integration of diffusion models into domain adaptation tasks. Diffusion models, originally proposed by \cite{sohl2015deep} and further developed by \cite{ho2020denoising}, have shown remarkable potential in bridging the gap between source and target domains. Diffusion models offer several advantages in domain adaptation tasks, including gradual transformation, which ensures smooth and controlled transitions between domains while preserving critical data features \cite{peng2024unsupervised}. Moreover, the ability of diffusion models to generate high-fidelity samples enables more accurate domain translation and feature adaptation \cite{benigmim2023one}.

This paper \cite{song2024stylegan} introduces StyleGAN-Fusion, a novel method that leverages pre-trained text-to-image diffusion models for domain adaptation of GAN generators without requiring ground truth samples from target domains. The key innovation is the use of classifier-free guidance from diffusion models as a critic to guide GANs in generating images aligned with text prompts. The method also introduces a diffusion directional regularizer to prevent model collapse and maintain diversity. This framework \cite{gao2023back} leverages diffusion models to adapt features from the source to the target domain by training a generation model on source data and using it to project target domain inputs back to the source domain for classification during inference. This paper \cite{wangbridging} introduces a novel method, DPMs-ANT (Diffusion Probabilistic Models with Adversarial Noise-Based Transfer Learning), to address the limited data problem in diffusion probabilistic models (DPMs) used for image generation. The method incorporates two key strategies: similarity-guided training which uses a classifier to estimate domain divergence, and adversarial noise selection which adaptively chooses noise based on the input image. 

This paper \cite{peng2024unsupervised} presents a novel method for unsupervised domain adaptation using diffusion models. It introduces the Domain-Adaptive Diffusion (DAD) module, which gradually transforms source domain features into the target domain while preserving semantic integrity. Alongside, a Mutual Learning Strategy (MLS) is utilized to facilitate bidirectional learning between the classification model and the DAD module, enhancing the model's adaptability through iterative feedback. The methods employed in this paper \cite{du2024diffusion} utilize a combination of diffusion-based models and active learning techniques for uncertainty estimation and domain adaptation. The approach centers around selecting informative target samples for annotation by applying a t-test-based selection strategy to address predictive uncertainties. The training process integrates the use of a diffusion classifier, which is trained on the target domain samples with a specific focus on modeling the probabilistic distribution of predictions.

\section{Diffusion-based Noise-centered Adversarial Learning Domain Adaptation}

\subsection{Problem formulation}

In the task of cross-user HAR domain adaptation, the term 'domain' refers to the 'user'. Domain adaptation, in this case, means transferring knowledge between different users. 

Given a labelled source user $\displaystyle S^{Source} =\left\{\left( x_{i}^{Source} ,\ y_{i}^{Source}\right)\right\}_{i=1}^{n^{Source}} $ drawn from a joint probability distribution $\displaystyle P^{Source} $ and a target user 

$\displaystyle S^{Target} =\left\{\left( x_{i}^{Target} ,\ y_{i}^{Target}\right)\right\}_{i=1}^{n^{Target}} $ drawn from a joint probability distribution $\displaystyle P^{Target} $, where $\displaystyle {n^{Source}} $ and $\displaystyle {n^{Target}} $ are the number of source and target samples respectively.  $\displaystyle S^{Source}$ and $\displaystyle S^{Target}$ have the same feature spaces (i.e. the set of features that describes the data from sensor readings) and label spaces (i.e. the set of activity classes). The source and target users have different distributions, i.e., $\displaystyle P^{Source} \neq P^{Target} $, which means that even for the same activity, the sensor readings look different between the two users. Given source user data $\displaystyle \left\{\left( x_{i}^{Source} ,\ y_{i}^{Source}\right)\right\}_{i=1}^{n^{Source}} $ and target user data $\displaystyle \left\{\left( x_{i}^{Target} \right)\right\}_{i=1}^{n^{Target}} $, the goal is to obtain the labels for the target user activities.

\subsection{An overview of the DNA-DA method}

Diffusion models, inspired by processes in physics, have recently shown great potential in various machine learning tasks. These models iteratively transform data from a simple distribution to a more complex one through a series of gradual steps, enabling the capture of intricate data structures. Diffusion models are particularly effective for domain adaptation due to their ability to learn complex transformations and smooth transitions between data distributions. By modeling the progressive changes needed to align source and target data distributions, diffusion models can effectively bridge the gap between different users' activity data.

\begin{figure}[h!]
\centering
\includegraphics[width=0.8\columnwidth]{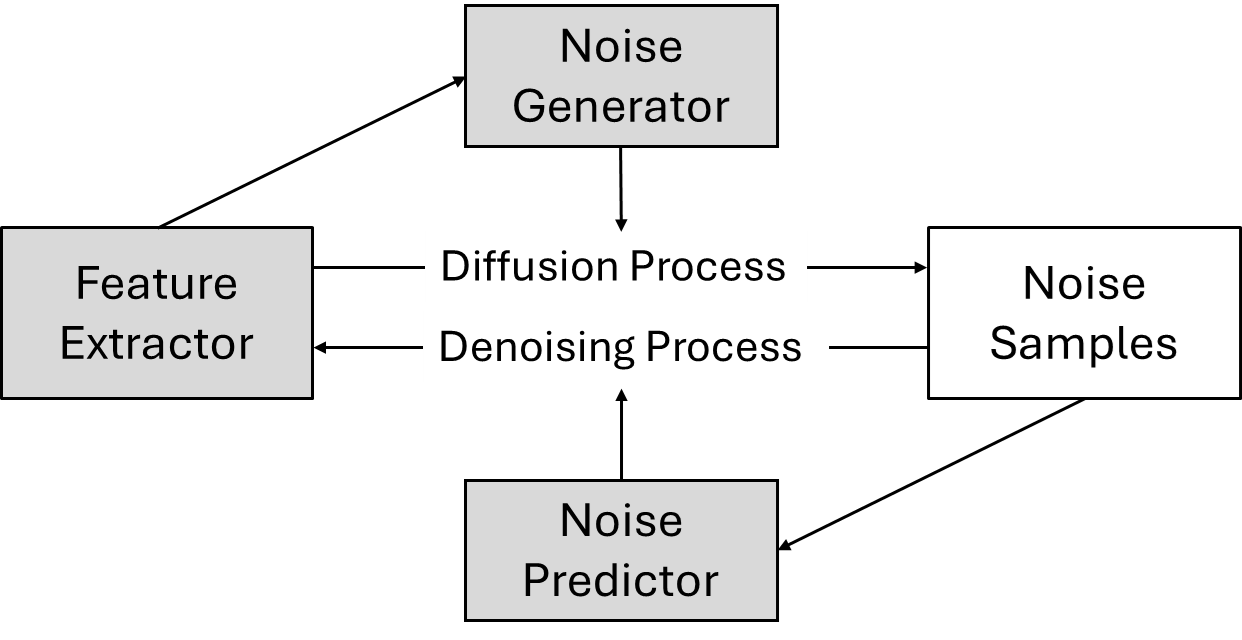}
\caption{The overview of the DNA-DA method.\label{DNA-DA-overview}}
\end{figure}

In the forward diffusion process, data is gradually transformed into noise that resembles random fluctuations. This noise, however, is often underutilized. By incorporating activity and domain class information into the noise, we make it follow various Gaussian distributions based on these conditional inputs, thereby enriching the noise with classification details. Therefore, we design a component of Noise Generator as illustrated in Figure~\ref{DNA-DA-overview}, to empower the new capability on noise. This component learns a projection from features to a Gaussian distribution, then the noise is sampled from this Gaussian distribution based on the activity classes and users label information. These noise is added to the feature that maps the feature to the direction of activity classes and users label.

During the reverse diffusion denoising process, this noise, now linked to classification features, carries significant information about the classification labels. By accurately predicting this reverse noise, the model can extract and use this embedded information effectively, as illustrated in Figure~\ref{DNA-DA-overview}. It is worthy to highligh that the noise predictor makes use of the the Gradient Reversal Layer (GRL) technique \cite{ganin2016domain}. The GRL adversarial learning method works by reversing the gradients during training. Its main objective is to ensure that the noise predicted by the model are not overly specific to individual users but instead contribute to a more generalized model. This approach helps the model create representations that are invariant to user-specific details. As a result, this technique is incorporated as a constraint in the denoising process to guide the model's learning process.

Moreover, both the forward diffusion and reverse denoising processes of DNA-DA method function as adversarial learning phases. In the forward diffusion phase, the goal is to differentiate between user and activity classes, acting as a discriminator to help the model recognize distinct characteristics of different users and their activities. In contrast, the reverse denoising phase aims to obscure the classification-related data distribution, acting as a generator. This adversarial setup helps the model better align feature distributions across different users, reducing discrepancies and improving generalization.

\subsection{Forward Diffusion Process}
In this section, we compare the origianl design of forward diffusion process of diffusion model to our novel design of DNA-DA.

\subsubsection{Original Forward Diffusion Process}

The forward diffusion process is designed to incrementally add noise to data over multiple timesteps. 

1. \textbf{Forward Diffusion Process Definition}:
\[
q(\mathbf{x}_t \mid \mathbf{x}_{t-1}) = \mathcal{N}(\mathbf{x}_t \mid {\alpha_t} \mathbf{x}_{t-1}, \beta_t \mathbf{I})
\]
where \(\beta_t = 1 - \alpha_t\) and \(\mathbf{I}\) represents the identity matrix.

This equation defines the forward diffusion process. It specifies that the distribution of \(\mathbf{x}_t\) given \(\mathbf{x}_{t-1}\) is a Gaussian distribution with mean \({\alpha_t} \mathbf{x}_{t-1}\) and variance \(\beta_t \mathbf{I}\). \(\alpha_t\) and \(\beta_t\) are parameters controlling the noise added at each time step. \(\beta_t = 1 - \alpha_t\) determines the variance of the noise. As \(t\) increases, \(\mathbf{x}_t\) becomes progressively noisier, transitioning from a signal to pure noise.

2. \textbf{Mean of Forward Process}:
\[
\mathbb{E}[\mathbf{x}_t \mid \mathbf{x}_{t-1}] = {\alpha_t} \mathbf{x}_{t-1}
\]

This equation provides the mean of the Gaussian distribution for \(\mathbf{x}_t\) given \(\mathbf{x}_{t-1}\). The mean is \({\alpha_t} \mathbf{x}_{t-1}\), showing that \(\mathbf{x}_t\) is a scaled version of \(\mathbf{x}_{t-1}\), with the scaling factor \({\alpha_t}\). The mean is scaled by \({\alpha_t}\), meaning the influence of \(\mathbf{x}_{t-1}\) on \(\mathbf{x}_t\) decreases as \(\alpha_t\) changes. This scaling factor controls how much of the previous state influences the current state.

3. \textbf{Variance of Forward Process}:
\[
\text{Var}[\mathbf{x}_t \mid \mathbf{x}_{t-1}] = \beta_t \mathbf{I}
\]

This equation provides the variance of the Gaussian distribution for \(\mathbf{x}_t\) given \(\mathbf{x}_{t-1}\). The variance represents the amount of noise added at each step. As \(t\) increases, \(\beta_t\) generally increases, making \(\mathbf{x}_t\) more noisy and less similar to \(\mathbf{x}_{t-1}\).

4. \textbf{Forward Noising Process}:
\[
\mathbf{x}_t = {\alpha_t} \mathbf{x}_{t-1} + {\beta_t} \mathbf{\epsilon}
\]
where \(\mathbf{\epsilon} \sim \mathcal{N}(\mathbf{0}, \mathbf{I})\) is the noise added to the process.
This equation describes how \(\mathbf{x}_t\) is generated from \(\mathbf{x}_{t-1}\) by adding noise. \(\mathbf{\epsilon}\) is drawn from a standard normal distribution \(\mathcal{N}(\mathbf{0}, \mathbf{I})\). \(\mathbf{x}_t\) is obtained by adding noise to a scaled version of \(\mathbf{x}_{t-1}\). \({\alpha_t} \mathbf{x}_{t-1}\) represents the signal component, and \({\beta_t} \mathbf{\epsilon}\) represents the noise component. As the process progresses, the noise component becomes more significant, making \(\mathbf{x}_t\) increasingly noisy and different from \(\mathbf{x}_{t-1}\).

\subsubsection{Forward Diffusion Process with Classification Information}

In our new design, we incorporate classification information directly into the forward diffusion process. This integration allows the model to utilize activity labels and user labels to better guide the orientation of noise addition process, empowering the noise as a carrier of the classification information.

1. \textbf{Forward Diffusion Process Definition}:
\[
q(\mathbf{x}_t \mid \mathbf{x}_{t-1}, \mathbf{c}_a, \mathbf{c}_u) = \mathcal{N}(\mathbf{x}_t \mid {\alpha_t} \mathbf{x}_{t-1} + \beta_t \left( \mathbf{c}_a \cdot \gamma_a + \mathbf{c}_u \cdot \gamma_u \right), \beta_t \mathbf{I})
\]
where \(\beta_t = 1 - \alpha_t\).

This equation defines the forward diffusion process with the inclusion of classification information. The distribution of \(\mathbf{x}_t\) given \(\mathbf{x}_{t-1}\), activity label \(\mathbf{c}_a\), and user label \(\mathbf{c}_u\) is a Gaussian distribution with a mean that combines the scaled previous data \({\alpha_t} \mathbf{x}_{t-1}\) and additional terms \(\mathbf{c}_a \cdot \gamma_a + \mathbf{c}_u \cdot \gamma_u\) based on the classification labels, and variance \(\beta_t \mathbf{I}\). \(\mathbf{c}_a\) and \(\mathbf{c}_u\) represent the activity and user labels, respectively. \(\gamma_a\) and \(\gamma_u\) are hyperparameters that control the influence of the activity and user labels on the data. The mean of the Gaussian distribution now incorporates these classification labels to guide the noising process. \(\beta_t = 1 - \alpha_t\) determines the amount of the noise added.

2. \textbf{Mean of Forward Process with Classification}:
\[
\mathbb{E}[\mathbf{x}_t \mid \mathbf{x}_{t-1}, \mathbf{c}_a, \mathbf{c}_u] = {\alpha_t} \mathbf{x}_{t-1} + \beta_t \left( \mathbf{c}_a \cdot \gamma_a + \mathbf{c}_u \cdot \gamma_u \right)
\]

This equation provides the mean of the Gaussian distribution for \(\mathbf{x}_t\) given \(\mathbf{x}_{t-1}\), activity label \(\mathbf{c}_a\), and user label \(\mathbf{c}_u\). The mean is a combination of the scaled previous data and the contributions from the classification labels. The mean now includes additional terms \(\mathbf{c}_a \cdot \gamma_a\) and \(\mathbf{c}_u \cdot \gamma_u\) that encode the influence of the activity and user labels. This incorporation helps align the data with the classification information.

3. \textbf{Variance of Forward Process}:
\[
\text{Var}[\mathbf{x}_t \mid \mathbf{x}_{t-1}, \mathbf{c}_a, \mathbf{c}_u] = \beta_t
\]

This equation provides the variance of the Gaussian distribution for \(\mathbf{x}_t\) given \(\mathbf{x}_{t-1}\), activity label \(\mathbf{c}_a\), and user label \(\mathbf{c}_u\). The variance remains \(\beta_t\), as it is independent of the classification labels. The variance of the noise is not affected by the classification information; it remains as \(\beta_t\), which governs the amount of noise added at each step.

4. \textbf{Forward Noising Process with Classification}:
\[
\mathbf{x}_t = {\alpha_t} \mathbf{x}_{t-1} + {\beta_t} \mathbf{\epsilon}
\]
where \(\mathbf{\epsilon} \sim \mathcal{N}( \mathbf{c}_a \cdot \gamma_a + \mathbf{c}_u \cdot \gamma_u, \mathbf{I})\) is the noise added to the process.

This equation describes the forward noising process with the integration of classification information. The data \(\mathbf{x}_t\) is generated from \(\mathbf{x}_{t-1}\) by adding noise with the information that incorporate activity and user labels. The term \({\alpha_t} \mathbf{x}_{t-1}\) represents the scaled previous data. \(\mathbf{c}_a \cdot \gamma_a\) and \(\mathbf{c}_u \cdot \gamma_u\) add classification-specific information to the noise. \({\beta_t} \mathbf{\epsilon}\) adds random noise to the process. This design improves how well the model aligns noise with classification labels.

\begin{figure}[h!]
\centering
\includegraphics[width=1\columnwidth]{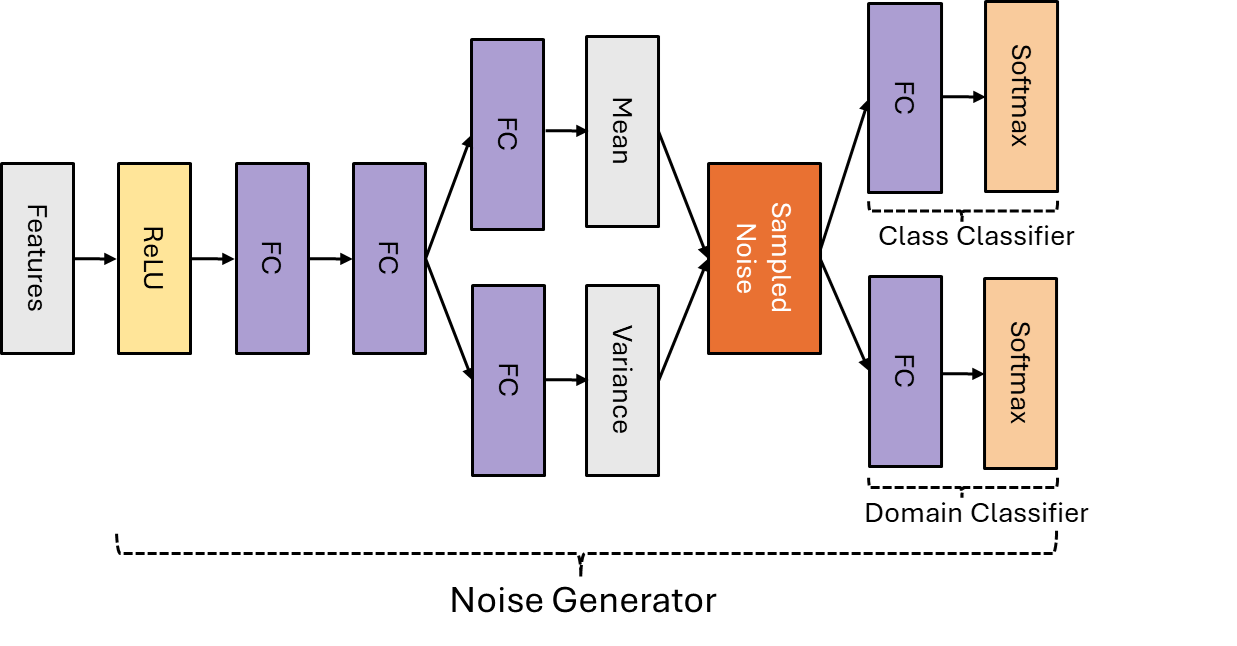}
\caption{The network architecture of the noise generator.\label{NG}}
\end{figure}

The network architecture of the noise generator is shown in Figure~\ref{NG}. This component learns to map input features to the mean and variance of a Gaussian distribution. Noise is then sampled from this distribution. The learning process is guided by both a class classifier and a domain classifier. The network begins with an input layer that takes in the extracted features. It includes several fully connected layers designed to learn complex patterns and transformations within the data. Notably, the ReLU (Rectified Linear Unit) activation function is used sparingly in this design. This is intentional, as we aim to maintain the linearity of the features with the added noise information, which helps regularize the model's classification ability. Following these layers, there are separate fully connected layers that output the parameters for the Gaussian distribution—specifically, the mean and variance. Noise sampled from this Gaussian distribution is a crucial part of our method, adding randomness to the generation process. The final layer employs two softmax activation functions, which is commonly used in classification tasks to produce a probability distribution over predicted output classes. This helps guide the orientation of the noise generation process, ensuring it aligns with the desired classification goals.

\subsection{Reverse Denoising Process}

In this section, we compare the origianl design of reverse denoising process of diffusion model to our novel design of DNA-DA.

\subsubsection{Original Reverse Denoising Process}

The original reverse denoising process aims to reconstruct the original data from the noisy data by learning the mean and variance parameters at each step. This process gradually removes the added noise and retrieves the underlying clean data.

1. \textbf{Reverse Process Definition}:
\[
p_\theta(\mathbf{x}_{t-1} \mid \mathbf{x}_t) = \mathcal{N}(\mathbf{x}_{t-1} \mid \tilde{\mu}_t(\mathbf{x}_t), \tilde{\beta}_t \mathbf{I})
\]
where \(\tilde{\mu}_t(\mathbf{x}_t)\) is the learned mean and \(\tilde{\beta}_t\) is the learned variance.

This equation defines the reverse denoising process. The distribution of \(\mathbf{x}_{t-1}\) given \(\mathbf{x}_t\) is a Gaussian distribution with mean \(\tilde{\mu}_t(\mathbf{x}_t)\) and variance \(\tilde{\beta}_t \mathbf{I}\). The reverse process aims to recover \(\mathbf{x}_{t-1}\) from the noisy \(\mathbf{x}_t\). \(\tilde{\mu}_t(\mathbf{x}_t)\) is the mean that is learned by the model, which represents the most probable value of \(\mathbf{x}_{t-1}\) given \(\mathbf{x}_t\). \(\tilde{\beta}_t\) is the variance that the model learns, indicating the uncertainty in the prediction of \(\mathbf{x}_{t-1}\).

2. \textbf{Learned Mean of Reverse Process}:
\[
\tilde{\mu}_t(\mathbf{x}_t) = \frac{1}{{\alpha_t}} \left( \mathbf{x}_t - \frac{\beta_t}{{1 - \bar{\alpha}_t}} \mathbf{\epsilon} \right)
\]
where \(\mathbf{\epsilon}\) is the noise term.

This equation provides the learned mean \(\tilde{\mu}_t(\mathbf{x}_t)\) for the reverse process. It is computed by adjusting \(\mathbf{x}_t\) based on the learned noise \(\mathbf{\epsilon}\) and the parameters \(\alpha_t\) and \(\beta_t\). \(\tilde{\mu}_t(\mathbf{x}_t)\) represents the estimate of the clean data \(\mathbf{x}_{t-1}\). The term \(\frac{1}{{\alpha_t}}\) scales \(\mathbf{x}_t\) to account for the signal contribution. The term \(\frac{\beta_t}{{1 - \bar{\alpha}_t}} \mathbf{\epsilon}\) adjusts for the noise added during the forward process. \(\mathbf{\epsilon}\) represents the noise term that is used to estimate the noise component in the reverse process.

3. \textbf{Learned Variance of Reverse Process}:
\[
\tilde{\beta}_t = \frac{\beta_t (1 - \bar{\alpha}_{t-1})}{1 - \bar{\alpha}_t}
\]

This equation provides the learned variance \(\tilde{\beta}_t\) for the reverse process. It is computed based on the variance \(\beta_t\) and the cumulative product of \(\alpha_t\) values, denoted \(\bar{\alpha}_t\). \(\tilde{\beta}_t\) represents the variance of the noise added during the reverse process. It is derived from the variance \(\beta_t\) used in the forward process and adjusted by the ratio of cumulative \(\alpha\) terms. This ensures that the noise level added during the reverse process is consistent with the noise level added during the forward process.

4. \textbf{Reverse Noising Process}:
\[
\mathbf{x}_{t-1} = \tilde{\mu}_t(\mathbf{x}_t) + \tilde{\beta}_t \mathbf{\eta}
\]
where \(\mathbf{\eta} \sim \mathcal{N}(\mathbf{0}, \mathbf{I})\) is the noise sampled in the reverse process.

This equation describes the reverse noising process. The clean data \(\mathbf{x}_{t-1}\) is obtained from \(\mathbf{x}_t\) by adding noise sampled from a standard normal distribution to the learned mean \(\tilde{\mu}_t(\mathbf{x}_t)\). \(\mathbf{x}_{t-1}\) is reconstructed by starting from the learned mean \(\tilde{\mu}_t(\mathbf{x}_t)\) and adding noise \(\tilde{\beta}_t \mathbf{\eta}\). \(\tilde{\beta}_t\) scales the noise term \(\mathbf{\eta}\) to account for the variance in the reverse process. \(\mathbf{\eta}\) is sampled from a standard normal distribution and represents the random noise added during the reverse process. This process iteratively denoises \(\mathbf{x}_t\) to recover the original clean data as \(\mathbf{x}_{t-1}\).

\subsubsection{Reverse Denoising Process with Adversarial Regularization}

In our new design, we enhance the reverse denoising process by incorporating adversarial regularization, which aims to improve the alignment between source and target distributions during the denoising process.

1. \textbf{Reverse Process Definition with Adversarial Regularization}:
\[
p_\theta(\mathbf{x}_{t-1} \mid \mathbf{x}_t, \mathbf{c}_a, \mathbf{c}_u, \mathbf{c}_{as}) = \mathcal{N}(\mathbf{x}_{t-1} \mid \tilde{\mu}_t(\mathbf{x}_t, \mathbf{c}_a, \mathbf{c}_u, \mathbf{c}_{as}), \tilde{\beta}_t \mathbf{I})
\]
where \(\tilde{\mu}_t(\mathbf{x}_t, \mathbf{c}_a, \mathbf{c}_u, \mathbf{c}_{as})\) incorporates adversarial regularization.

This equation defines the reverse denoising process with adversarial regularization. The distribution of \(\mathbf{x}_{t-1}\) given \(\mathbf{x}_t\), activity label \(\mathbf{c}_a\), user label \(\mathbf{c}_u\), and source user activity label \(\mathbf{c}_{as}\) is a Gaussian distribution with a learned mean \(\tilde{\mu}_t(\mathbf{x}_t, \mathbf{c}_a, \mathbf{c}_u, \mathbf{c}_{as})\) and variance \(\tilde{\beta}_t \mathbf{I}\). The reverse process aims to recover \(\mathbf{x}_{t-1}\) from the noisy \(\mathbf{x}_t\). The learned mean \(\tilde{\mu}_t\) now incorporates adversarial regularization to ensure better alignment between source and target distributions. \(\tilde{\beta}_t\) is the learned variance, representing the uncertainty in the prediction of \(\mathbf{x}_{t-1}\).

2. \textbf{Learned Mean of Reverse Process with Adversarial Regularization}:

\begin{equation}
\begin{split}
\tilde{\mu}_t(\mathbf{x}_t, \mathbf{c}_a, \mathbf{c}_u, \mathbf{c}_{as}) &= \frac{1}{\alpha_t} \left( \mathbf{x}_t - \frac{\beta_t}{1 - \bar{\alpha}_t} \left( \text{AR}(\mathbf{c}_a \cdot \gamma_a + \mathbf{c}_u \cdot \gamma_u) \right. \right. \\
&\left. \left. + \mathbf{c}_{as} \cdot \gamma_{as} \right) \right)
\end{split}
\end{equation}

where \(\text{AR}(\cdot)\) denotes the adversarial mechanism of Gradient Reversal technique applied to \(\mathbf{c}_a \cdot \gamma_a + \mathbf{c}_u \cdot \gamma_u\).

This equation provides the learned mean \(\tilde{\mu}_t\) for the reverse process, incorporating adversarial regularization. The mean is computed by adjusting \(\mathbf{x}_t\) based on the learned noise and adversarial terms. The adversarial regularization \(\text{AR}(\cdot)\) is applied to the classification information to ensure feature user-invariant. \(\tilde{\mu}_t(\mathbf{x}_t, \mathbf{c}_a, \mathbf{c}_u, \mathbf{c}_{as})\) represents the estimate of the clean data \(\mathbf{x}_{t-1}\) with adversarial adjustments. The term \(\frac{1}{{\alpha_t}}\) scales \(\mathbf{x}_t\) to account for the signal contribution. The term \(\frac{\beta_t}{{1 - \bar{\alpha}_t}}\) adjusts for the noise component. \(\text{AR}(\mathbf{c}_a \cdot \gamma_a + \mathbf{c}_u \cdot \gamma_u)\) applies adversarial regularization to ensure that the activity and user labels do not bias the learned mean. \(\mathbf{c}_{as} \cdot \gamma_{as}\) represents the adversarial signal related to the source user activity label, helping to align features between the source and target domains.

3. \textbf{Learned Variance of Reverse Process}:
\[
\tilde{\beta}_t = \frac{\beta_t (1 - \bar{\alpha}_{t-1})}{1 - \bar{\alpha}_t}
\]

This equation provides the learned variance \(\tilde{\beta}_t\) for the reverse process. It is computed based on the variance \(\beta_t\) and the cumulative product of \(\alpha\) values, denoted \(\bar{\alpha}_t\). \(\tilde{\beta}_t\) represents the variance of the noise added during the reverse process. It is derived from the forward process variance \(\beta_t\) and adjusted by the ratio of cumulative \(\alpha\) terms. This ensures that the noise level added during the reverse process is consistent with the forward process, while incorporating adversarial adjustments if needed.

4. \textbf{Reverse Noising Process with Adversarial Regularization}:
\[
\mathbf{x}_{t-1} = \tilde{\mu}_t(\mathbf{x}_t, \mathbf{c}_a, \mathbf{c}_u, \mathbf{c}_{as}) + \tilde{\beta}_t \mathbf{\eta}
\]
where \(\mathbf{\eta} \sim \mathcal{N}(\mathbf{0}, \mathbf{I})\) is the noise sampled in the reverse process.

This equation describes the reverse noising process with adversarial regularization. The clean data \(\mathbf{x}_{t-1}\) is reconstructed from \(\mathbf{x}_t\) by adding noise sampled from a standard normal distribution to the learned mean \(\tilde{\mu}_t\). \(\mathbf{x}_{t-1}\) is obtained by starting from the learned mean \(\tilde{\mu}_t\) and adding noise \(\tilde{\beta}_t \mathbf{\eta}\). \(\tilde{\beta}_t\) scales the noise term \(\mathbf{\eta}\) to account for the variance in the reverse process. \(\mathbf{\eta}\) is sampled from a standard normal distribution and represents the random noise added during the reverse process.  Adversarial regularization helps align features between different users, improving the quality of the reconstructed data and ensuring consistency with the source and target distributions.

By incorporating adversarial regularization terms into the learned mean, the reverse denoising process is guided to focus on minimizing the discrepancy between source and target distributions. This approach leverages adversarial learning to enhance the robustness of the reverse process and improve the final reconstructed data quality.

\begin{figure}[h!]
\centering
\includegraphics[width=1.0\columnwidth]{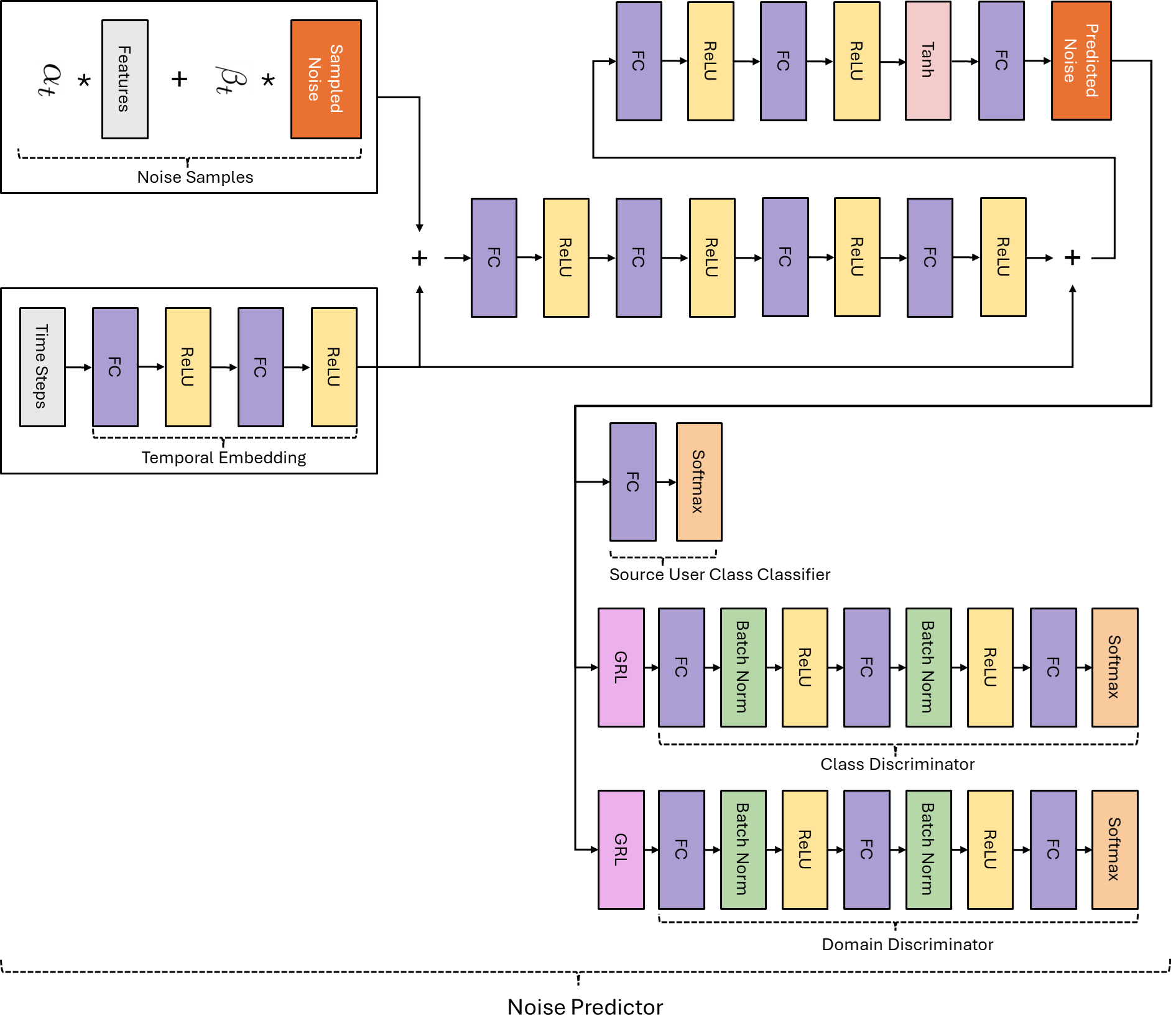}
\caption{The network architecture of the noise predictor.\label{NP}}
\end{figure}

The network architecture of the noise predictor is shown in Figure~\ref{NP}. The Noise Samples are the combination of features and sampled noise from the forward diffusion process. The Temporal Embedding is used to encode time steps to provide the temporal information to the noise samples, where samples is gradually reconstructed into noise. Additionally, the design of temporal embedding is inspired by the residual networks \cite{he2016identity}. This approach helps address the issue of forgetting temporal information in deep networks. Then, these fully connected layers are the foundation of the neural network, responsible for learning and representing complex patterns from the input data. Moreover, the learning of the Predicted Noise is guided by the source user class classifier during the training process, while the activity class discriminator and user domain discriminator are learned in the reverse gradient adversarial training way to confuse the model for learning a user-invariant activity classifier. The discriminator has a more complex structure compared to the classifier because it is designed to learn more intricate mappings. This increased complexity allows the discriminator to effectively confuse the model by distinguishing between subtle differences, improving overall model generalization.

In this way, the forward diffusion and reverse denoising processes cooperate together and function as adversarial learning domain adaptation. During the forward diffusion phase, the focus is on distinguishing between different user and activity classes, acting as a discriminator to help the model identify the unique features of various users and their activities. Conversely, the reverse denoising phase aims to obscure the data distribution related to classification, functioning as a generator. This adversarial framework helps the model better align feature distributions across different users, reducing discrepancies and enhancing its ability to generalize.

\subsection{Training Objectives}

\subsubsection{Original Training Objectives}

The original training objective for diffusion models is to maximize the data likelihood by minimizing the reconstruction error between the original data and the denoised data. This objective aims to ensure that the learned mean and variance parameters effectively reverse the noise addition process.

\textbf{Noise Prediction Objective}:
\[
\mathcal{L}_{\text{total}} = \mathcal{L}_{\text{noise}} = \mathbb{E}_{\mathbf{x}_0, \mathbf{\epsilon}, t} \left[ \left\| \mathbf{\epsilon} - \hat{\mathbf{\epsilon}}(\mathbf{x}_t, t) \right\|^2 \right]
\]
where \(\hat{\mathbf{\epsilon}}(\mathbf{x}_t, t)\) is the predicted noise from the model.

This objective function measures the difference between the true noise \(\mathbf{\epsilon}\) and the predicted noise \(\hat{\mathbf{\epsilon}}(\mathbf{x}_t, t)\) at a given timestep \(t\). The goal is to minimize this difference. The expectation \(\mathbb{E}_{\mathbf{x}_0, \mathbf{\epsilon}, t}\) is taken over the original data \(\mathbf{x}_0\), the noise \(\mathbf{\epsilon}\), and the time step \(t\). The term \(\left\| \mathbf{\epsilon} - \hat{\mathbf{\epsilon}}(\mathbf{x}_t, t) \right\|^2\) represents the squared error between the true noise and the predicted noise. By minimizing \(\mathcal{L}_{\text{noise}}\), the model learns to accurately predict the noise added during the forward diffusion process. Accurate noise prediction ensures that the model can effectively reverse the noise addition during the reverse process, recovering the original data.

\subsubsection{New Training Objectives}

In our new design, we introduce an adversarial regularization term to the training objective. This term encourages the model to generate denoised data that aligns more closely with the target distribution.

1. \textbf{Noise Prediction Objective}:
\[
\mathcal{L}_{\text{noise}} = \mathbb{E}_{\mathbf{x}_0, \mathbf{\epsilon}, t} \left[ \left\| \mathbf{\epsilon} - \hat{\mathbf{\epsilon}}(\mathbf{x}_t, t) \right\|^2 \right]
\]
where \(\hat{\mathbf{\epsilon}}(\mathbf{x}_t, t)\) is the predicted noise from the model.

This objective function measures the difference between the actual noise \(\mathbf{\epsilon}\) and the predicted noise \(\hat{\mathbf{\epsilon}}(\mathbf{x}_t, t)\) at time \(t\). The expectation \(\mathbb{E}_{\mathbf{x}_0, \mathbf{\epsilon}, t}\) is taken over the original data \(\mathbf{x}_0\), the noise \(\mathbf{\epsilon}\), and the time step \(t\). The term \(\left\| \mathbf{\epsilon} - \hat{\mathbf{\epsilon}}(\mathbf{x}_t, t) \right\|^2\) represents the squared error between the true noise and the model's prediction. Minimizing \(\mathcal{L}_{\text{noise}}\) ensures that the model learns to accurately predict the noise, which is crucial for effectively reversing the forward diffusion process.

2. \textbf{Classification Objectives in Forward Diffusion Process}:

   Activity Classification (\(\mathbf{c}_a\)):
   \[
   \mathcal{L}_{\text{act}} = \mathbb{E}_{\mathbf{x}_0, \mathbf{c}_a} \left[ \text{CrossEntropy}(\hat{\mathbf{c}}_a, \mathbf{c}_a) \right]
   \]
   
   User Classification (\(\mathbf{c}_u\)):
   \[
   \mathcal{L}_{\text{binary}} = \mathbb{E}_{\mathbf{x}_0, \mathbf{c}_u} \left[ \text{CrossEntropy}(\hat{\mathbf{c}}_u, \mathbf{c}_u) \right]
   \]

These objectives guide the model to classify activity and domain labels during the forward diffusion process. \(\mathcal{L}_{\text{act}}\) ensures that the model can accurately predict the activity label \(\mathbf{c}_a\) from the data. For this classifier, we only have access to labels from the source users, while the labels for target users are not available. Even though we don't know the specific classes in the target user, it is often assumed that the relationship between samples and their classes is understood \cite{uslu2022segmentation}. For instance, we might know a sample belongs to a certain class, say class A, but we don’t know what class A represents (e.g., walking, running, etc.). Therefore, we can assign arbitrary numbers to each class in the target user, but these numbers will differ from the labels used in the source user. For more detailed implementation, refer to the paper \cite{hallac2017toeplitz}. On the other side, \(\mathcal{L}_{\text{binary}}\) focuses on predicting the domain classification \(\mathbf{c}_u\), which is a binary classifier that distinguish source and target users. The users' binary labels are known. Both terms use cross-entropy loss to measure the discrepancy between the predicted labels \(\hat{\mathbf{c}}_a\) and \(\hat{\mathbf{c}}_u\) and the true labels \(\mathbf{c}_a\) and \(\mathbf{c}_u\), respectively.

3. \textbf{Classification Objectives in Reverse Diffusion Process}:

   Adversarial Learning Loss:
\[
\mathcal{L}_{\text{adv}_a} = -\mathbb{E}_{\mathbf{x}_0, \mathbf{c}_a, \mathbf{c}_u} \left[ \text{CrossEntropy}(\hat{\mathbf{c}}_a, \mathbf{c}_a) \right]
\]

\[
\mathcal{L}_{\text{adv}_u} = -\mathbb{E}_{\mathbf{x}_0, \mathbf{c}_a, \mathbf{c}_u} \left[ \text{CrossEntropy}(\hat{\mathbf{c}}_u, \mathbf{c}_u) \right]
\]

\[
\mathcal{L}_{\text{adv}} = \mathcal{L}_{\text{adv}_a} + \mathcal{L}_{\text{adv}_u}
\]

   Activity Classification of Source User (\(\mathbf{c}_{as}\)):
   \[
   \mathcal{L}_{\text{act-source}} = \mathbb{E}_{\mathbf{x}_0, \mathbf{c}_{as}} \left[ \text{CrossEntropy}(\hat{\mathbf{c}}_{as}, \mathbf{c}_{as}) \right]
   \]

These objectives introduce adversarial regularization and source user activity classification into the reverse denoising process. \(\mathcal{L}_{\text{adv}}\) applies adversarial regularization to ensure that the model's predictions for activity and domain classifications align with the target distribution. It uses cross-entropy loss and aims to minimize the classification accuracy, encouraging the model to generalize better across domains. \(\mathcal{L}_{\text{act-source}}\) focuses on classifying the activity label of the source user, helping to maintain the relevance of source-specific information in the reverse process.

4. \textbf{Combined Training Objective}:
\begin{equation}
\begin{split}
\mathcal{L}_{\text{total}} &= \mathcal{L}_{\text{noise}} + \lambda_{\text{act}} \mathcal{L}_{\text{act}} + \lambda_{\text{binary}} \mathcal{L}_{\text{binary}} \\
&\quad + \lambda_{\text{adv}} \mathcal{L}_{\text{adv}} + \lambda_{\text{act-source}} \mathcal{L}_{\text{act-source}}
\end{split}
\end{equation}
where \(\lambda_{\text{act}}\), \(\lambda_{\text{binary}}\), \(\lambda_{\text{adv}}\), and \(\lambda_{\text{act-source}}\) are hyperparameters that control the contribution of each term.

The combined training objective integrates the noise prediction loss with classification and adversarial losses, weighted by their respective hyperparameters. \(\mathcal{L}_{\text{total}}\) balances multiple objectives to guide the model's training. \(\mathcal{L}_{\text{noise}}\) ensures effective noise prediction. \(\mathcal{L}_{\text{act}}\) and \(\mathcal{L}_{\text{binary}}\) ensure accurate classification during the forward process. \(\mathcal{L}_{\text{adv}}\) enforces adversarial regularization to improve generalization across users. \(\mathcal{L}_{\text{act-source}}\) maintains the relevance of source user activity information. Hyperparameters \(\lambda_{\text{act}}\), \(\lambda_{\text{binary}}\), \(\lambda_{\text{adv}}\), and \(\lambda_{\text{act-source}}\) are tuned to balance the trade-offs between different training objectives. The incorporation of adversarial learning into the training objective aims to improve the model's ability to generalize across different users by maximising and minimizing the discrepancy between source and target distributions. This approach leverages the strengths of both diffusion models and adversarial learning to achieve robust cross-user HAR domain adaptation.

\section{Experiments}

In this section, we evaluate the performance of our DNA-DA method through a series of comprehensive experiments focused on cross-user HAR.

\subsection{Datasets and Experimental Setup}

For sensor-based HAR research, OPPT, PAMAP2, and DSADS are the most representative and commonly used. We selected these datasets to cover a broad range of daily and sports-related activities to validate our cross-user HAR approach. Details about the subjects and activities common across these datasets are shown in Table~\ref{tab_datasets_info}. Our focus is on IMU sensor (i.e. accelerometer and gyroscope) data collected from the lower right arm, simulating a smartwatch-based scenario.

\begin{table*}[ht!]
\caption{Information on the Sensor-based HAR Datasets.}
\label{tab_datasets_info}
\centering
\resizebox{\textwidth}{!}{%
\begin{tabular}{|l|l|l|}
\hline
\textbf{Dataset} & \textbf{Subjects} & \textbf{Common Activities} \\ \hline
OPPT & S1, S2, S3 & 1 standing, 2 walking, 3 sitting, 4 lying \\ \hline
PAMAP2 & 1, 5, 6 & \begin{tabular}[c]{@{}l@{}}1 lying, 2 sitting, 3 standing, 4 walking,\\ 5 running, 6 cycling, 7 Nordic walking,\\ 8 ascending stairs, 9 descending stairs,\\ 10 vacuum cleaning, 11 ironing\end{tabular} \\ \hline
DSADS & 2, 4, 7 & \begin{tabular}[c]{@{}l@{}}1 sitting, 2 standing, 3 lying on back,\\ 4 lying on right, 5 ascending stairs,\\ 6 descending stairs, 7 standing in elevator still,\\ 8 moving around in elevator, 9 walking in parking lot,\\ 10 walking on treadmill in flat, 11 walking on treadmill inclined positions,\\ 12 running on treadmill in flat, 13 exercising on stepper,\\ 14 exercising on cross trainer, 15 cycling on exercise bike in horizontal positions,\\ 16 cycling on exercise bike in vertical positions, 17 rowing, 18 jumping,\\ 19 playing basketball\end{tabular} \\ \hline
\end{tabular}%
}
\end{table*}

The OPPORTUNITY (OPPT) dataset \cite{chavarriaga2013opportunity} records subjects performing daily activities with minimal restrictions, following only a general description of tasks. The PAMAP2 dataset \cite{reiss2012introducing} features a more structured protocol with over 10 hours of data per participant, detailing specific activities for each person. Lastly, the Daily and Sports Activities Data Set (DSADS) \cite{barshan2014recognizing} involves participants performing activities in their own way, leading to high variability that reflects real-world conditions. Each activity is performed for 5 minutes per participant.

These datasets are selected based on increasing levels of complexity for domain adaptation: OPPT with 4 activities, PAMAP2 with 11 activities, and DSADS with 19 activities. Notably, DSADS includes closely related activities like different types of walking, making it more challenging to classify.

For our experiments, we use a sliding window approach to segment the data, a common technique in sensor-based HAR \cite{wang2018impact}. Each window spans 3 seconds with a 50\% overlap, capturing temporal relationships. We have approximately 20,000 sensor samples per activity for OPPT and PAMAP2, and 7,500 samples per activity for DSADS.

In the training phase, we use labeled data from a source user and unlabeled data from a target user. The target user's data is split into a validation set and a test set: the first half serves as the validation set for tuning hyperparameters, while the second half is used for testing \cite{xiaozhou2023temporaloptimal}. This temporal partitioning simulates new data being introduced over time. We select hyperparameters based on the validation set to ensure the model generalizes well and does not overfit. Finally, we evaluate the model's performance on the test set using classification accuracy, assessing how effectively the model can classify activities for the target user.

\subsection{Data Distribution Analysis}

Delving deeper into the analysis, we explore the cross-user data distribution differences of activities across the three datasets. We apply a common feature extraction method in HAR \cite{xiaozhou2023temporaloptimal} to sliding windows for both source and target users in each dataset. To quantify the differences in feature distributions between users, we employ the Wasserstein distance \cite{vallender1974calculation}. This metric measures the effort required to transform one distribution into another, considering the cost of moving and transforming distribution mass, which aligns well with our case.

The observed distance represents the Wasserstein distance of the complete data of the feature distributions in each activity between source and target users, depicted as the red dashed lines. This metric gauges the degree of $\displaystyle o.o.d.$ of an activity across users, where a higher value indicates greater dissimilarity between the two data distributions. Next, we conduct resampling with replacement from each distribution to generate new sample distributions, repeating this process 5000 times. The bootstrap distance represents the Wasserstein distance of the resampled feature distributions in each activity between source and target users, visualized as the blue distributions. Bootstrapped distances aid in understanding the variability and centrality of the distance measure, revealing how different the feature distributions of the sub-activities are between source and target users. If more bootstrap distances are higher than the observed distance, it indicates greater data distribution difference of the subset of the activity features, implying that the sub-activity feature distributions are more distant.

Furthermore, we introduce the Bootstrapped proportion, which denotes the proportion of distances from the bootstrapped samples that are smaller than or equal to the observed distance from the original samples. This proportion serves to measure the degree of sub-activity data distribution difference between users. A lower Bootstrapped proportion suggests that the sub-activity data distribution difference between users is more significant.

\begin{figure}[h!]
\centering
\includegraphics[width=1\columnwidth]{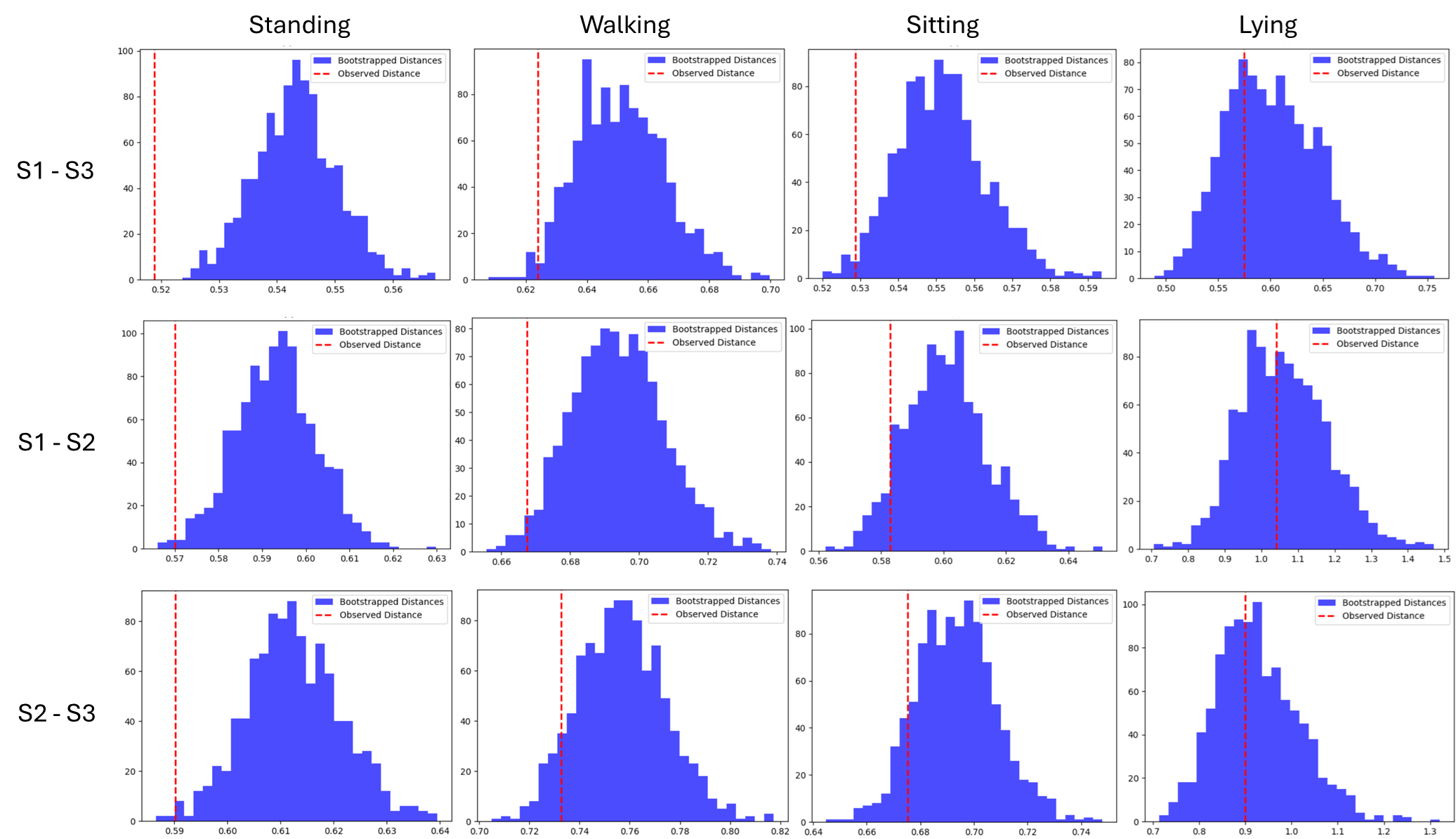}
\caption{Cross-user data distribution distance of activities in OPPT dataset. \label{OPPT_DD_Compare}}
\end{figure}

In Figure~\ref{OPPT_DD_Compare}, we observe that the feature distributions of the lying activity across users differ significantly compared to standing, as indicated by the varying distances from the red dashed lines. However, when examining the sub-activity feature distributions, represented by the blue part, the situation is reversed. This suggests that while the lying activity remains relatively consistent, there is more variability in the standing activity. The reason for this disparity could be attributed to the consistent sleep posture across users during the lying activity, whereas individuals tend to change their posture more frequently while standing.

\begin{figure}[h!]
\centering
\includegraphics[width=1\columnwidth]{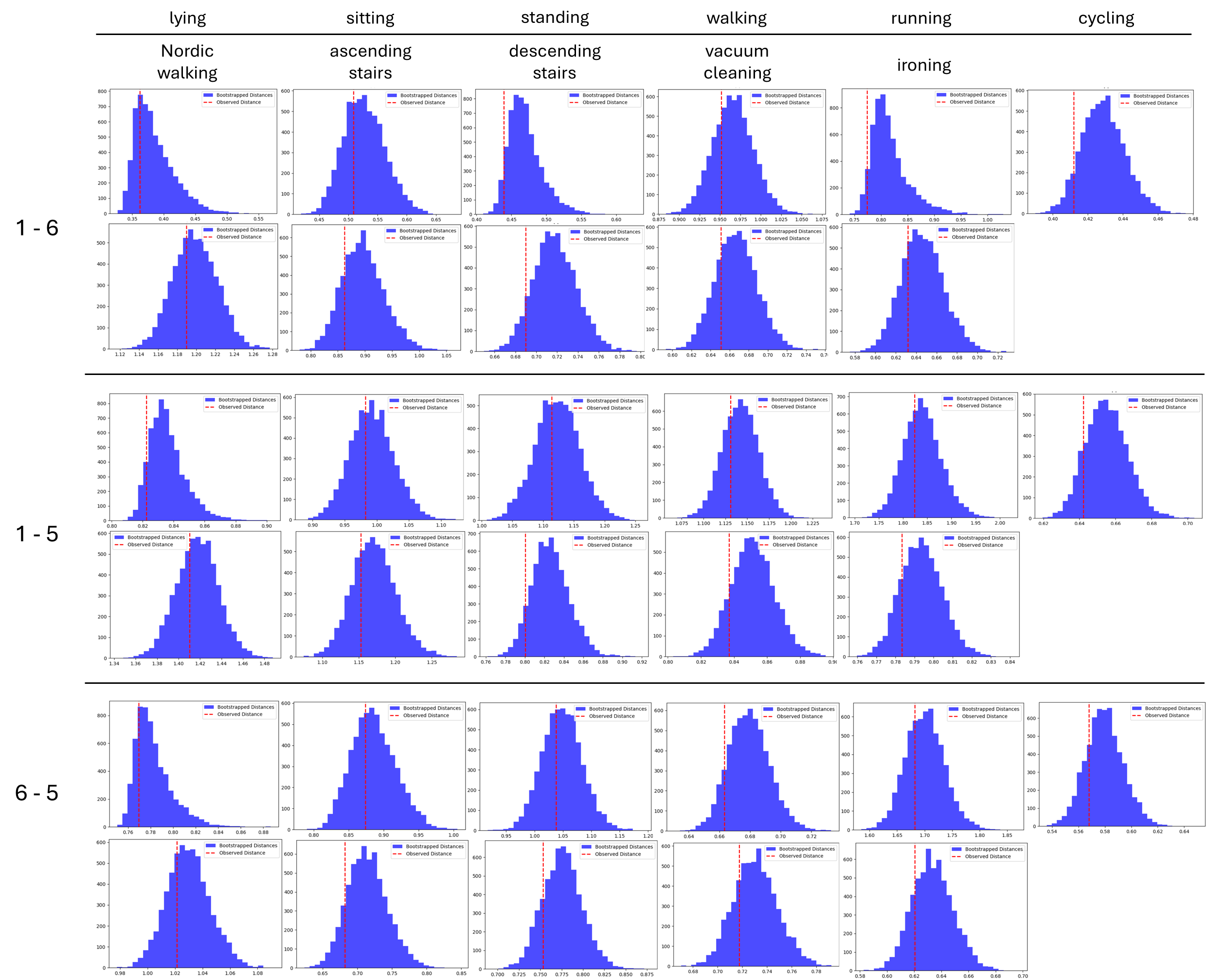}
\caption{Cross-user data distribution distance of activities in PAMAP2 dataset. \label{PAMAP2_DD_Compare}}
\end{figure}

In Figure~\ref{PAMAP2_DD_Compare}, we note that the feature distribution distance for Nordic walking activity exceeds 1.02 across all three users, indicating significant variability. Conversely, the feature distribution distance for cycling remains consistently below 0.65 across all users. For the running and walking activity, we observe a feature distribution distance exceeding 1.65 from user 5 to users 1 and 6. However, users 1 and 6 exhibit a relatively smaller feature distribution distance, suggesting more similar activities between them. Similarly, for ascending and descending stairs activities, the feature distribution distance is small from users 6 to user 1 and 5. However, users 1 and 5 show a relatively larger feature distribution distance, indicating a distinct activity pattern. In the sub-activity feature distribution view, we observe greater variability in ascending and descending stairs activities, likely due to the continuous change in positions

\begin{figure*}[h!]
\centering
\includegraphics[width=1\textwidth]{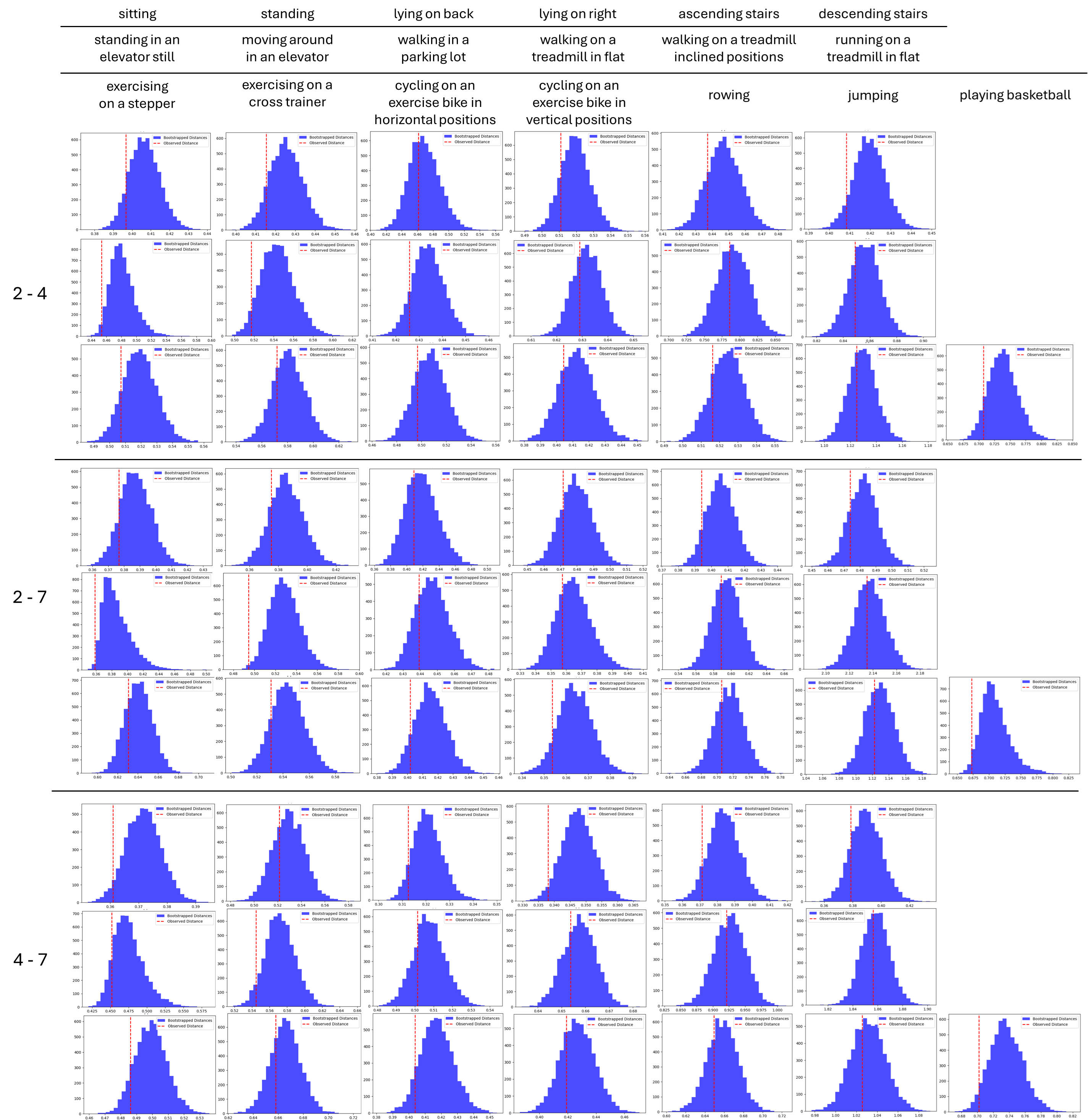}
\caption{Cross-user data distribution distance of activities in DSADS dataset. \label{DSADS_DD_Compare}}
\end{figure*}

In Figure~\ref{DSADS_DD_Compare}, we observe that the feature distribution distance for the jumping activity exceeds 1.02 across all three users, indicating significant variability. Conversely, the feature distribution for sitting remains consistently low across all users. Activities such as playing basketball and moving around in an elevator exhibit high differences in sub-activity feature distributions, likely due to varying activity habits among users. However, activities like running on a treadmill in a flat position and walking on a treadmill at an inclined position show minimal variability in sub-activity features, suggesting that the features follow similar distributions for each user.

\begin{table*}[ht!]
\centering
\caption{Comparison of the average feature distributions distance of Activities and the average Bootstrapped proportion of Activities between users on three datasets}
\label{tab:dataset_comparison_distance_pvalue}
\begin{tabular}{@{}lccc@{}}
\toprule
\textbf{Dataset} & \textbf{Users} & \textbf{Avg. Distance} & \textbf{Avg. BS Prop.} \\ \midrule
\multirow{3}{*}{OPPT} & S1 - S2 & 0.715 & 0.140 \\
                      & S1 - S3 & 0.561 & 0.082 \\
                      & S2 - S3 & 0.724 & 0.155 \\ \midrule
\multirow{3}{*}{PAMAP2} & 1 - 5 & 1.045 & 0.247 \\
                        & 1 - 6 & 0.679 & 0.193 \\
                        & 5 - 6 & 0.854 & 0.238 \\ \midrule
\multirow{3}{*}{DSADS} & 2 - 4 & 0.559 & 0.188 \\
                       & 2 - 7 & 0.594 & 0.194 \\
                       & 4 - 7 & 0.608 & 0.190 \\ \bottomrule
\end{tabular}
\end{table*}

In Table~\ref{tab:dataset_comparison_distance_pvalue}, we analyze the differences in data distribution from both a holistic user and dataset perspective. The 'average distance' represents the average observed distances of all activities, while the 'average Bootstrapped proportion' signifies the average Bootstrapped proportion of activities. Our findings indicate that the OPPT and PAMAP2 datasets exhibit more significant differences in feature distribution between activities compared to the DSADS dataset. This suggests a greater variation in activity features within OPPT and PAMAP2. Moreover, when examining sub-activity data distribution differences between users, we observe more pronounced disparities within the OPPT and DSADS datasets. This suggests that users in these datasets engage in sub-activities in more distinct ways compared to those in the PAMAP2 dataset. These distinct sub-activities reflect different data distributions, providing suitable cross-user datasets for our subsequent research endeavors.

\subsection{Results and Discussion}

The experimental analysis aims to evaluate the performance of various domain adaptation methods on the three datasets, focusing on user adaptation tasks. Specifically, we compare six methods of CORAL, SOT, DANN, OTDA, Raincoat and SPA to our DNA-DA.

\begin{itemize}
\item \textbf{CORAL} \cite{sun2016return}: CORAL (Covariate Shift Adaptation) adjusts the covariance matrices of feature layers to improve feature alignment across different domains, helping features become more domain-agnostic.

\item \textbf{SOT} \cite{lu2021cross}: The SOT (Substructure Optimal Transport) method explores the internal structure of domains to enable precise mapping at a substructure level. It effectively balances general and specific mapping requirements.

\item \textbf{DANN} \cite{ganin2016domain}: DANN (Domain-Adversarial Neural Network) uses adversarial training to make a neural network’s features indistinguishable between source and target domains. The approach includes a discriminator that learns to identify whether features come from the source or target domain, while the network learns to produce features that fool the discriminator.

\item \textbf{OTDA} \cite{flamary2016optimal}: OTDA (Optimal Transport Domain Adaptation) employs optimal transport theory to efficiently map one probability distribution onto another, helping align samples from different domains.

\item \textbf{Raincoat} \cite{he2023domain}: Raincoat addresses domain adaptation in time series by managing both feature and label shifts. It aligns features and labels across domains by integrating time and frequency information, and corrects any discrepancies between domains.

\item \textbf{SPA} \cite{xiao2024spa}: The SPA (Spectral Alignment) method focuses on both inter-domain transferability and intra-domain structure. SPA transforms the domain adaptation problem into graph-based primitives, using a coarse graph alignment approach combined with a spectral regularizer to align domain graphs in eigenspaces.
\end{itemize}

\begin{table}[t]
\centering
\begin{tabular}{|l|l|}
\hline
\textbf{Parameter} & \textbf{Value} \\
\hline
Training Epochs & 500 \\
Learning Rate & 0.001 \\
$ \lambda_{\text{act}}$,$\lambda_{\text{binary}}$,$\lambda_{\text{adv}}$,$\lambda_{\text{act-source}}$  & 1.0 \\
$\gamma_a$,$\gamma_u$  & 1.0 \\
\hline
\end{tabular}
\caption{Parameter settings of the DNA-DA Method.\label{DNA-DA_paras}}
\end{table}

The primary metric for assessment is the classification accuracy for the target user. For the proposed DNA-DA method, parameter settings are listed in Table~\ref{DNA-DA_paras}. The different coefficients of constraint loss can be set based on the importance of these constraints. In our experiments setting, we unify the coefficients to be one to simplified the hyperparameters tuning process. In a similar way, we also unify the coefficients that control the influence of the activity and user labels on the data during forward diffusion process to be one. Better experiment results can be achieved by more fine-grained tuning of these coefficients.

From the experimental results, it is evident that DNA-DA significantly outperforms all other methods across all the tasks. Below is a detailed analysis of each method's performance:

\begin{figure}[h!]
\centering
\includegraphics[width=\columnwidth]{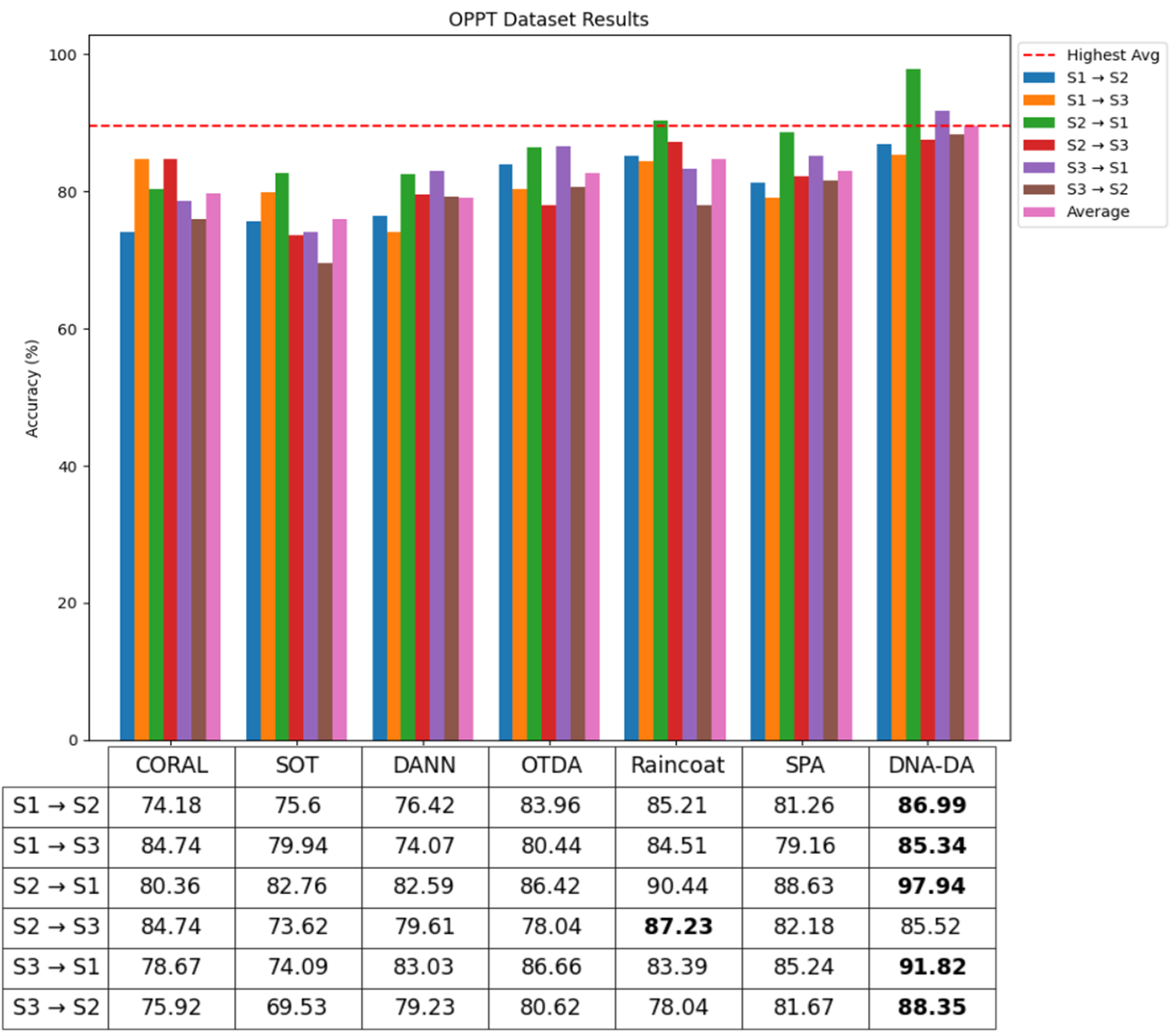}
\caption{OPPT dataset classification results.\label{OPPT_Dataset_Results}}
\end{figure}

On the OPPT dataset (see Figure~\ref{OPPT_Dataset_Results}), DNA-DA stands out with exceptional performance, achieving the highest accuracy across all scenarios. DNA-DA's accuracy peaks at 97.94\% in the S2 → S1 scenario, showcasing its superior adaptability and robustness in cross-subject activity recognition. Raincoat and TrC also perform commendably, with Raincoat achieving its highest accuracy of 90.44\% in the S2 → S1 scenario, and TrC showing a peak accuracy of 86.99\% in the S1 → S2 scenario. In contrast, other methods like OTDA and SA exhibit more moderate performance, with accuracies ranging from mid-70\% to mid-80\%. CORAL and SOT generally show lower performance, with accuracies below 80\%, indicating their limitations in effectively managing the variability and complexity of the OPPT dataset.

\begin{figure}[h!]
\centering
\includegraphics[width=\columnwidth]{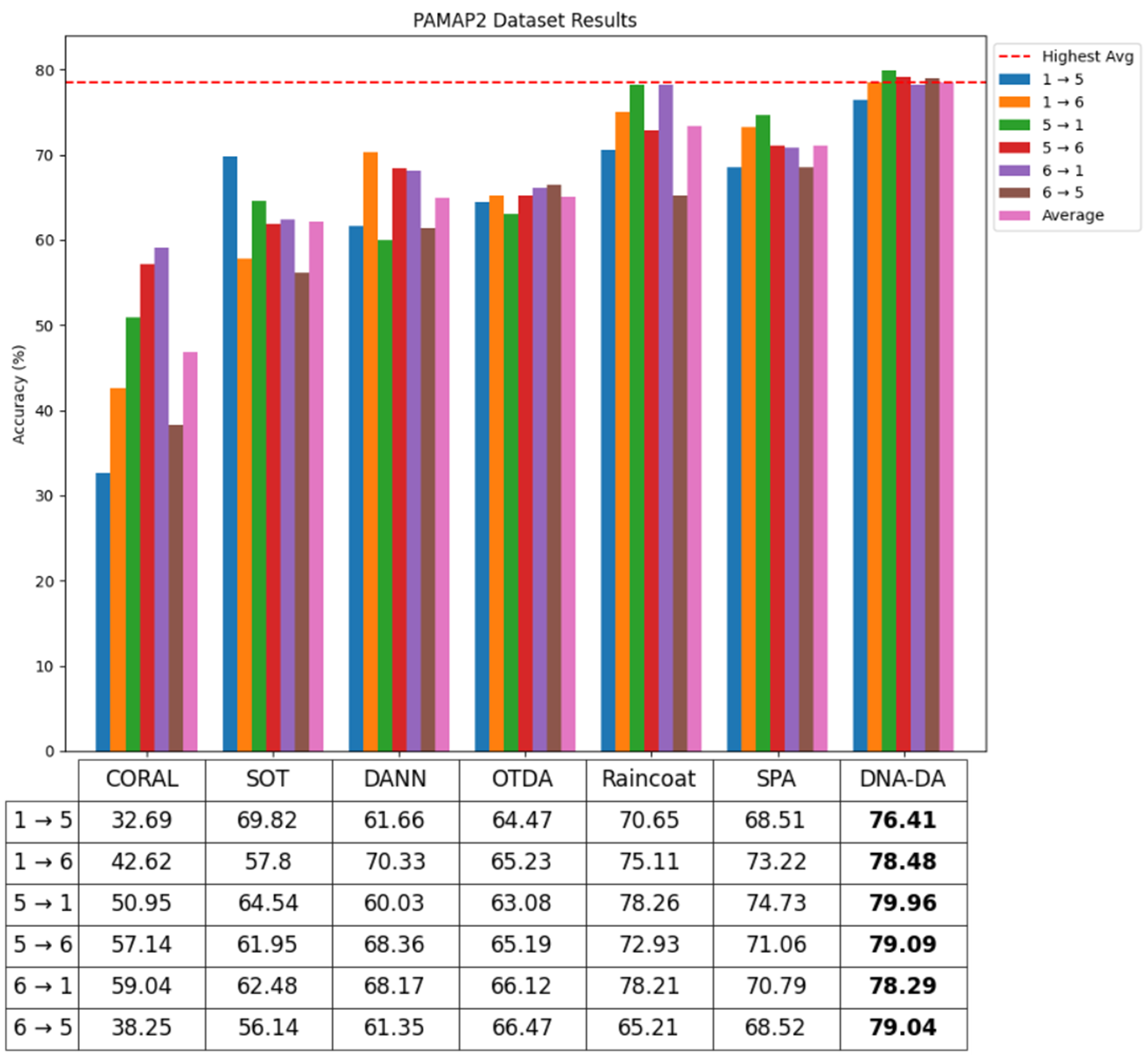}
\caption{PAMAP2 dataset classification results.\label{PAMAP2_Dataset_Results}}
\end{figure}

For the PAMAP2 dataset (see Figure~\ref{PAMAP2_Dataset_Results}), DNA-DA again demonstrates its robustness, with accuracy consistently above 76\% and peaking at 79.96\% in the 5 → 1 scenario. This highlights DNA-DA's effective adaptation to the diverse set of activities in the PAMAP2 dataset. Raincoat also shows strong performance, especially in the 5 → 1 scenario (78.26\%) and the 1 → 6 scenario (75.11\%). TrC achieves its highest accuracy of 79.96\% in the 5 → 6 scenario, but its overall performance is less consistent compared to DNA-DA. Methods such as DANN and OTDA display respectable accuracy in the range of 60\% to 70\%, while CORAL and SOT perform less effectively, with accuracies below 70\%. This indicates that CORAL and SOT struggle to adapt to the structured and varied conditions present in the PAMAP2 dataset.

\begin{figure}[h!]
\centering
\includegraphics[width=\columnwidth]{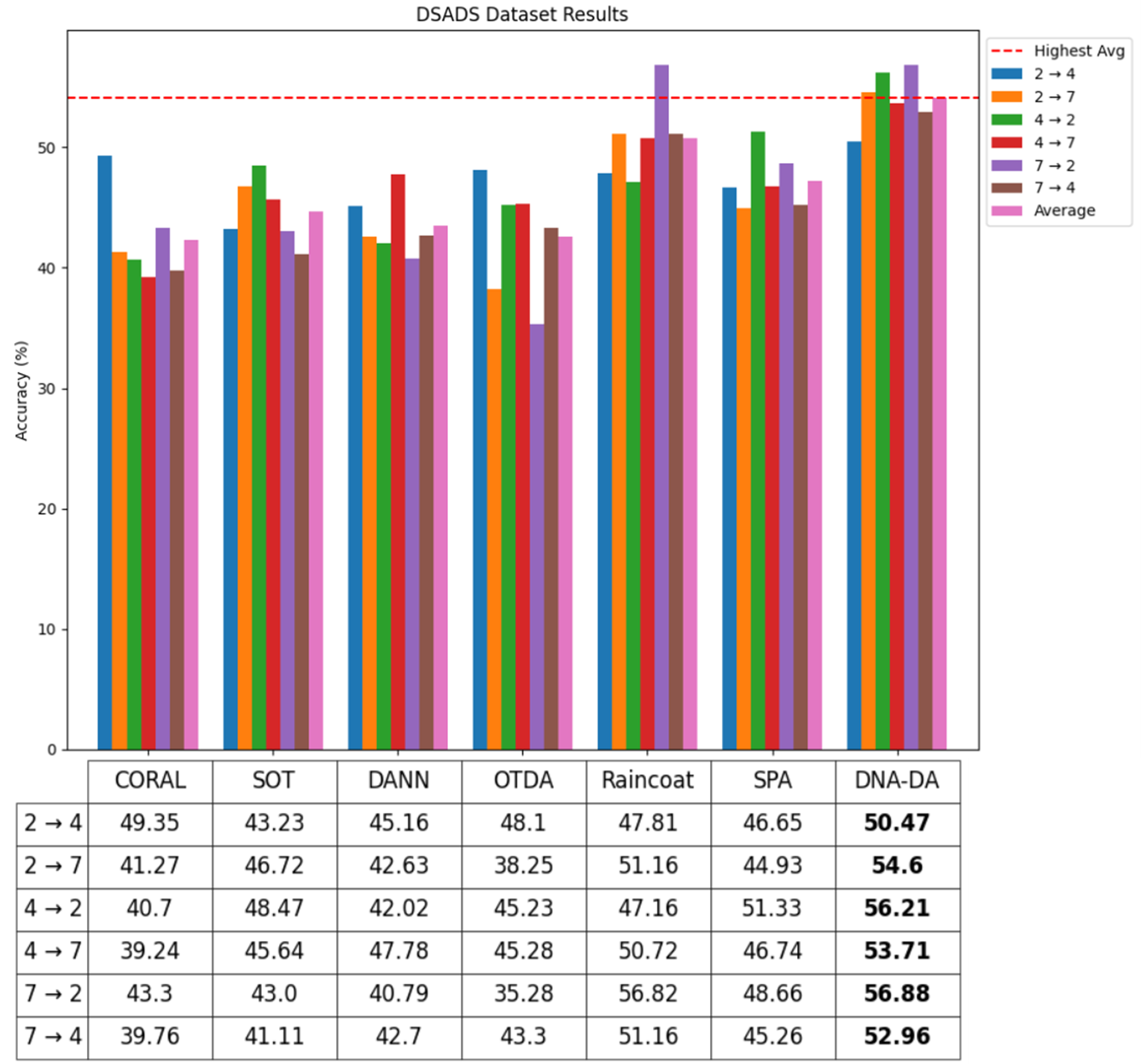}
\caption{DSADS dataset classification results.\label{DSADS_Dataset_Results}}
\end{figure}

In the DSADS dataset (see Figure~\ref{DSADS_Dataset_Results}), characterized by high variability in activities, DNA-DA emerges as the top performer with the highest accuracy across all scenarios, peaking at 56.88\% in the 7 → 2 scenario. This performance underscores DNA-DA's ability to handle the dataset's diverse and real-world conditions effectively. Raincoat and TrC also show reasonable performance, with Raincoat achieving up to 56.82\% and TrC reaching 56.21\%. Conversely, methods like DANN, OTDA, and SA exhibit lower performance, with accuracy in the range of 40\% to 50\%, highlighting their challenges in adapting to the variability and complexity of the DSADS dataset.

Overall, DNA-DA consistently proves to be the most effective method across all datasets, demonstrating superior adaptability and robustness in handling cross-user HAR tasks. DNA-DA’s ability to maintain high performance across different datasets underscores its robustness and generalizability. In comparison, Raincoat and TrC also show strong performance but are less consistent, with Raincoat performing particularly well in certain scenarios and TrC exhibiting variable results. Methods such as CORAL, SOT, and DANN display mixed results and generally perform less effectively, suggesting the need for further refinement to enhance their performance in diverse HAR contexts. The trends observed across these datasets highlight the reliability of DNA-DA and indicate potential areas for improvement in other methods. The experiments results demonstrate that the proposed cross-user data distribution alignment techniques DNA-DA significantly improve the generalization performance of HAR models.

\section{Conclusion}
In this paper, we introduced DNA-DA, a novel method for cross-user HAR domain adaptation. Our approach utilizes diffusion models combined with adversarial learning for domain adaptation. Specifically, we enhance the forward diffusion process by incorporating classification information into the noise, and we improve the reverse denoising process through adversarial regularization. The training objectives are set to achieve a max-min adversarial learning goal, which helps in better aligning source and target users' data distributions. Experimental results demonstrate the effectiveness of our method in achieving robust cross-user HAR domain adaptation, showcasing its potential for real-world applications.

For future work, we plan to further test the DNA-DA method with activities that have more distinct data distributions between users. This will help us evaluate the method's versatility and robustness in handling more challenging tasks. Additionally, we intend to broaden our experimental approach by including a wider range of user groups, such as individuals with different physical characteristics and those from different hand usage habits(i.e. left-handed or right-handed activities). This will help us determine how well DNA-DA performs across various segments of the population.

To strengthen our evaluation, we are developing our own datasets that cover a wide range of real-world scenarios. These custom datasets will provide insights into how well DNA-DA adapts to different situations beyond the controlled lab environment. This effort aims not only to validate our approach but also to contribute new and varied datasets to the research community.

\bibliographystyle{IEEEtran}
\bibliography{ref}

\end{document}